\documentclass[preprint,12pt]{elsarticle}

\usepackage{graphicx}
\usepackage{todonotes}

\usepackage[T1]{fontenc}

\usepackage{natbib}
\usepackage{color}

\usepackage[english]{babel}
\usepackage{url}
\usepackage[utf8]{inputenc}
\usepackage{amsmath, natbib}
\usepackage{array,multirow,graphicx}
\usepackage{rotating}
\usepackage{hyperref}

\begin{document}

\begin{frontmatter}

\title{Semantic Localization in the PCL library}

\author[alc,alb]{Jesus Mart\'inez-G\'omez}
\author[alc]{Vicente Morell}
\author[alc]{\corref{cor1}Miguel Cazorla}
\cortext[cor1]{Corresponding author}
\ead{miguel.cazorla@ua.es}
\author[alb]{Ismael Garc\'ia-Varea}
\address[alc]{Dpto. of Computer Science and Artificial Intelligence, University of Alicante., P.O. Box 99. 03080, Alicante, Spain.}
\address[alb]{Computer System Department, University of Castilla-La Mancha, Spain.}

\begin{abstract}
The semantic localization problem in robotics consists in determining the place where a robot is located by means of semantic categories. The problem is usually addressed as a supervised classification process, where input data correspond to robot perceptions while classes to semantic categories, like kitchen or corridor. 

In this paper we propose a framework, implemented in the PCL library, which provides a set of valuable tools to easily develop and evaluate semantic localization systems. The implementation includes the generation of 3D global descriptors following a Bag-of-Words approach. This allows the generation of dimensionality-fixed descriptors from any type of keypoint detector and feature extractor combinations. The framework has been designed, structured and implemented in order to be easily extended with different keypoint detectors, feature extractors as well as classification models.

The proposed framework has also been used to evaluate the performance of a set of already implemented descriptors, when used as input for a specific semantic localization system. The results obtained are discussed paying special attention to the internal parameters of the BoW descriptor generation process. Moreover, we also review the combination of some keypoint detectors with different 3D descriptor generation techniques. 

\end{abstract}


\end{frontmatter}

\section{Introduction}

The semantic localization problem can be defined as the problem of determining the place where a robot is located by means of semantic categories. The problem is usually addressed as a supervised classification process, where input data correspond to robot perceptions, and classes to semantic room/place categories, like kitchen, bathroom, or corridor. Commonly, this classification process is tackled by using models that require dimensionality-fixed inputs, such as SVMs~\cite{orabona2007ipr} or Bayesian Network classifiers~\cite{yi2009brl}. In order to transform robot perception into dimensionality-fixed descriptors, we can opt by using global features or build them from a set of local features following the well-known Bag-of-Words (BoW) approach~\cite{yang2007ebr}.

During the last decade, the semantic location problem has attracted the attention of the scientific community, becoming one of the well-known problems in robotics. In fact, several image processing techniques, evaluation datasets, open challenges, and different approaches has been proposed so far, as it is shown in a very recent published survey paper~\cite{kostavelis2015smf}. Actually, the semantic information about the place where the robot is located can be very helpful for more specific robotic tasks like autonomous navigation, high-level planning, simultaneous location and mapping (SLAM), or human-robot interaction.

The Point Cloud Library (PCL~\cite{rusu2011pcl}) has become, in less than four years from its first release, the most widely used open source project for 2D/3D image and point cloud processing. The PCL proposes several algorithms for most of the well-known problems in computer vision: feature extraction, surface reconstruction, image registration, model fitting, and segmentation. Moreover, it implements standard machine learning techniques for clustering and supervised classification. However, PCL does not currently provide a standard procedure for generating 3D global descriptors from local ones. This could be carried out by following a BoW approach, which would allow PCL users to take advantage of all the useful 3D local features included in the library for a wider range of problems. Concretely, any type of 3D local feature could be properly used as input for the semantic localization problem. 

In this article, we propose a PCL implementation of the BoW approach relying on machine learning techniques already implemented in the library. Several 3D global descriptors generated with such approach are evaluated when serving as input for the semantic localization problem. Therefore, the purpose of this work is two fold: in one hand to propose a general framework to easily develop and evaluate semantic localization systems using 3D point cloud information as input data; and on the other hand to implement it in the PCL, taking advantage of the availability of 3D image processing techniques. Both with the aim at providing a set of tools to be useful for the PCL community.

Then, the three major contributions of this work are: 

\begin{itemize}

\item The generation of 3D global descriptors from PCL local features following a Bag-of-Words approach, which will allow the generation of dimensionality-fixed descriptors from any kind of keypoint detector and feature extractor combination.

\item The definition of a common framework to develop and evaluate semantic localization systems within PCL. This framework has been designed and implemented to be easily extended with different and new keypoint detectors, feature extractors and classification models.

\item The experimentation carried out with a challenging benchmark, which provides sequences of labeled RGB-D images acquired with a mobile robot indoor office environments. In this experimentation, we evaluate the internal parameters that take part in the BoW approach (e.g. the dictionary size), but we also discuss the role of the keypoint detectors and feature extractors.

\end{itemize}

The rest of the paper is organized as follow: in Section~\ref{sec:semantic-localization}, a more detailed description of the semantic localization problem is presented, as well as a review of some recent proposal to deal with that problem. Section~\ref{sec:methodology} presents the design and development of the proposed framework. In Section~\ref{sec:pcl-contributions}, the specific contributions of this work to the PCL are described. In Section~\ref{sec:experiments} the experimental results carried out to demonstrate the functionality and usability of this work are presented. Finally, in Section~\ref{sec:conclussions} the main conclusions and future works are outlined.

\section{Semantic Localization\label{sec:semantic-localization}}

\subsection{Problem definition}

As stated before, the semantic localization problem can be formulated as a classical statistical pattern recognition problem as follows. Let $I$ be a perception from a robot (in our case an RGB-D image), $d(I)$ a function that generates a specific descriptor given $I$, and $M$ a classification model that provides the class posterior probability $P_M(c|d(I))$, where $c$ is a class label from a set of predefined class categories ${\cal C}$. Then, this problem can be stated, without loss of generality, as the problem of finding the optimal label $\hat{c}$ according to:
$$\hat{c} = \arg\max_{c\in{\cal C}} P_M(c|d(I))$$

In general, and following that approach, we can identify two main steps to be performed when designing and building a semantic localization system:
\begin{enumerate}

\item To carry out a descriptor generation process given the input perception.
\item To design a classifier capable of discriminating among the different types of scenes. This classifier will be trained using the descriptors generated in the previous step.
\end{enumerate}
A more detailed description of this two steps is shown in Section~\ref{sec:methodology}.

\subsection{Related work}

For a complete review of the state-of-art in semantic localization we refer the reader to~\cite{kostavelis2015smf} where a survey on this subject has been recently published. However, let's review the most related previous works from the last recent years.

As already mentioned, the semantic localization problem consists of the process of acquiring an image, generate a suitable representation (that is, an image descriptor) and classifying the imaged scene~\cite{wu2009vpc}. This classification can be performed according to a) high-level features of the environment, like detected objects \cite{ranganathan2007smo,vasudevan2008bsc,espinace2013isr}, b) global image representations~\cite{oliva2006btg}, or c) local features~\cite{tuytelaars2008lif}. In~\cite{torralba2003cvs} a method for scene classification based on global image features was presented, where the temporal continuity between consecutive images was exploited using a Hidden Markov Model. In~\cite{mozos2005slo}, a scene classifier with range data as input information and AdaBoost as the classification model is proposed. In 2006, Pronobis et al.~\cite{pronobis2006ada} developed a visual scene classifier using composed receptive field histograms~\cite{linde2004oru} and SVMs.

The use of the Bag of Words (BoW) technique~\cite{csurka2004vcw} can also be considered a remarkable milestone for visual semantic scene classification. The BoW process starts by creating a visual dictionary of representative features. Next, each extracted feature is assigned to the closest word in the dictionary. Then, a histogram representing the number of occurrences of each visual word is computed. This histogram is finally used as the image descriptor. An extensive evaluation of BoW features representations for scene classification were presented in~\cite{yang2007ebr}, demonstrating that visual words representations are likely to produce superior performance. In~\cite{lazebnik2006bbf}, an extension of the BoW technique using a spatial pyramid was proposed. Also, this work is one of the most relevant articles related to scene classification allowing to merge local and global information into a single image descriptor. The spatial pyramid approach has been successfully applied to several semantic localization problems, and it can be considered a standard solution for generating descriptors.

All mentioned works used visual cameras as input devices. However, visual cameras are highly affected by changing lighting conditions. The lighting variations can occur due to different external weather conditions, but also because of the presence or lack of artificial lights. This reason makes the use of RGB-D cameras very useful in current semantic localization approaches, even to deal with real-time constraints as proposed in~\cite{lim2012trs}.

\section{Framework Design\label{sec:methodology}}

In this section, we describe the BoW framework proposed to manage the semantic localization problem, which has been previously defined as a classical supervised classification problem. Therefore, we assume the following initial setup. We are provided with, at least, two sequences of RGB-D images acquired with a mobile robot. The RGB-D images represent scenes from an office indoor environment, such as Universities or Government buildings. Each RGB-D image from the first sequence (training) is labeled with the semantic category of the room where it was acquired, using labels as "kitchen" or "corridor". The problem consists in determining the label for the RGB-D images from the second sequence (test).
\\
The framework proposed includes the following steps:

\begin{enumerate}

\item Extract features from training and test RGB-D data. The goal of this step is to find an appropriate image representation, suitable for serving as input in subsequent steps. It involves a set of sub-tasks.

\begin{enumerate}

	\item Select a keypoint detection method, which reduces the amount of points to work with and speeds up the process. 
    
    \item Select a feature extraction procedure. The combination of keypoints and features should present some specific characteristics: efficiency, repeatability, distinctiveness and accuracy~\cite{tuytelaars2008lif,martinez2014taxonomy}.
    
    \item For each keypoint detected, extract the descriptor associated to the selected feature when possible. We can find some keypoints not meeting the features requirements, such as a number of surrounding points within a neighborhood. This fact can reduce the final number of features extracted from the RGB-D image.

\end{enumerate}

\item Transform the features extracted into global descriptors with fixed-dimensionality using a BoW approach.

\begin{enumerate}

	\item Merge all the features extracted from the complete training sequence into a single set of features.

	\item Perform a $k$-means clustering over this set to select a subset of $k$ representative features. This subset of features is known as the dictionary, and its size $k$ should have been previously defined.
    
    \item For each training and test RGB-D image, assign all their (previously extracted) features with the closest word in the dictionary. Then, compute a histogram over these assignations whose dimensionality corresponds to the dictionary size. This histogram is then used as image descriptor.

\end{enumerate}

\item Train a classification model using the training sequence. Based on the training descriptors generated in the previous step (and the room labels), we train a SVM classifier~\cite{vapnik:2000}. Thanks to the use of dimensionality-fixed inputs, most of the classifiers capable of managing continuous data could be used.

\item Classify the whole test sequence. The last step classifies each test descriptor with the SVM model computed in the training stage. 

\end{enumerate}

\begin{figure}[!tbh]
 \centering
\includegraphics[width=0.98\linewidth]{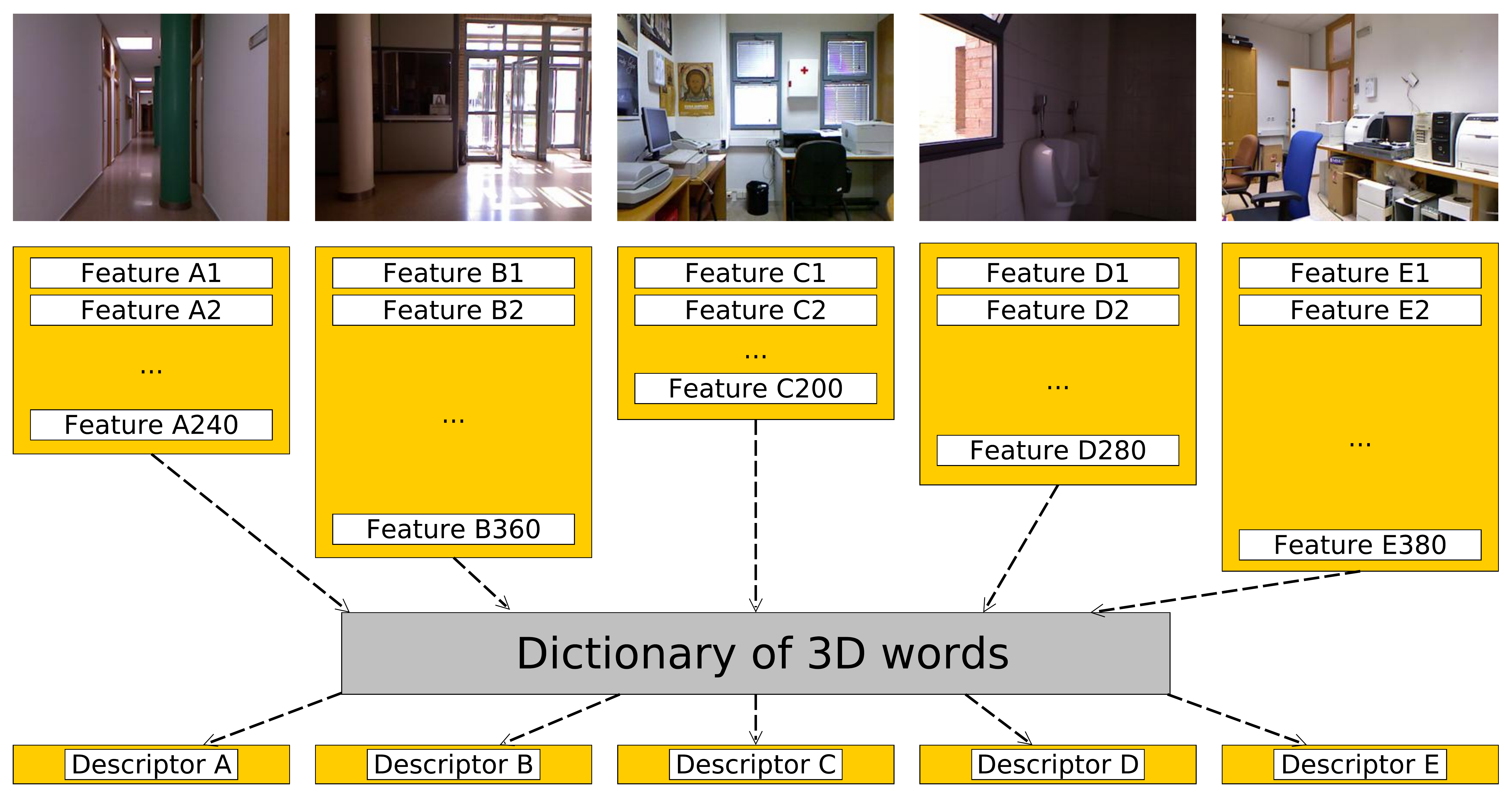}
 \caption{Descriptor generation process from the features extracted and a dictionary of 3D words previously computed.}
 \label{fig:fig1}
\end{figure}

Fig.~\ref{fig:fig1} shows the descriptor generation process from a set of features extracted. It can be observed how the final descriptor presents the same dimensionality for all the input images, even when a different number of features were extracted from them. 

\section{Point Cloud Library Contributions\label{sec:pcl-contributions}}

In this section, we describe the two main contributions for the PCL. The source code of the provided tool is available online under the Creative Commons Attribution license (CC-BY 3.0) at 
\begin{center}

   \href{https://bitbucket.org/vmorell/semanticlocalization}{https://bitbucket.org/vmorell/semanticlocalization}
\end{center}

\subsection{3D global descriptors from local features}

Although there are several global descriptors for 3D data, as previously commented in Section~\ref{sec:semantic-localization}, the BoW method could be used for describing the whole point cloud using local features. Local features could come from a combination of 3D keypoint detectors and features. This global feature, a histogram, could be used for other tasks purposes. In the presented framework, it is quite easy to modify the code to include different keypoint detectors and feature methods. We provide in the code some experimentation with some 3D local keypoint detectors and feature descriptors available in the PCL. We briefly describe them.

One of the simplest detector is Uniform Sampling (US). US builds a 3D voxel grid with the input data and takes the centroid (average point inside a voxel) of the voxel grid as keypoint. The resulting point cloud is then reduced and downsampled in a uniform way. Another keypoint detector is Harris3D~\cite{sipiran2011}. The implementation available in PCL takes the normals to the input pointcloud as the input for this detector. For each point, it selects points in a given neighborhood and calculates a covariance matrix of the normals at those points. Then, a value is calculated for each point based on the determinant and trace of the covariance matrix (as proposed in~\cite{Harris88acombined} for 2D). After a local maximum suppression method is applied, the surviving points are the keypoints for the input point cloud. 

The Normal Aligned Radial Feature (NARF)~\cite{steder2010narf} keypoint detector and feature descriptor use the range image to calculate the descriptor, not the point cloud. The keypoint detector find borders in the range image and calculates a score, indicating how the surface changes on each point. After this score is calculated, a smoothing process and non-maximum suppression are applied. With regard to the feature extraction process, NARF extracts a descriptor from each keypoint and its neighborhood. A star pattern is used, and for each beam of the pattern, it calculates the intensity changes along the cells lying under the beam. Then, for each beam, a value in the range $[-0.5, 0.5]$ is obtained. To make it invariant against rotation, the predominant orientation is calculated. Another feature used in the framework is the Signature of Histograms of OrienTations (SHOT)~\cite{tombari2010unique}. The descriptor is calculated by concatenating a set of local histograms over the 3D volume defined by a 3D grid centered at a keypoint. For each local histogram and for each point, the angular difference between the normal in the point and the normal in the keypoint is accumulated in the histogram. A variant is the Color-SHOT~\cite{colorSHOT2011} which adds a color histogram to the original SHOT descriptor.

Another two features used in our experiments are based on the Point Feature Histogram (PFH)~\cite{rusu2008aligning}. PFH selects from a keypoint, a set of points in a given neighborhood. For each two points in that neighborhood, PFH calculates four values which together express geometric relationship between those points. The four values are concatenated and a histogram is calculated using the values of all the possible combination of points. The first variation of the PFH is the Fast PFH~\cite{fpfh2009}, which improves the efficiency of the original PFH, not processing some points in the neighborhood. The second one is the PFH-RGB, which includes color to the geometrical information.

\subsection{Framework for semantic localization}

Our main contribution in this paper is the development of a framework that could be used for experimentation in semantic localization. Our main goal building this framework is the suitability for future development, i.e., it must be easy to integrate different keypoint detectors and feature descriptors, as well as to use others classification methods.

\begin{figure}[!htb]
\centering
\includegraphics[width=0.9\linewidth]{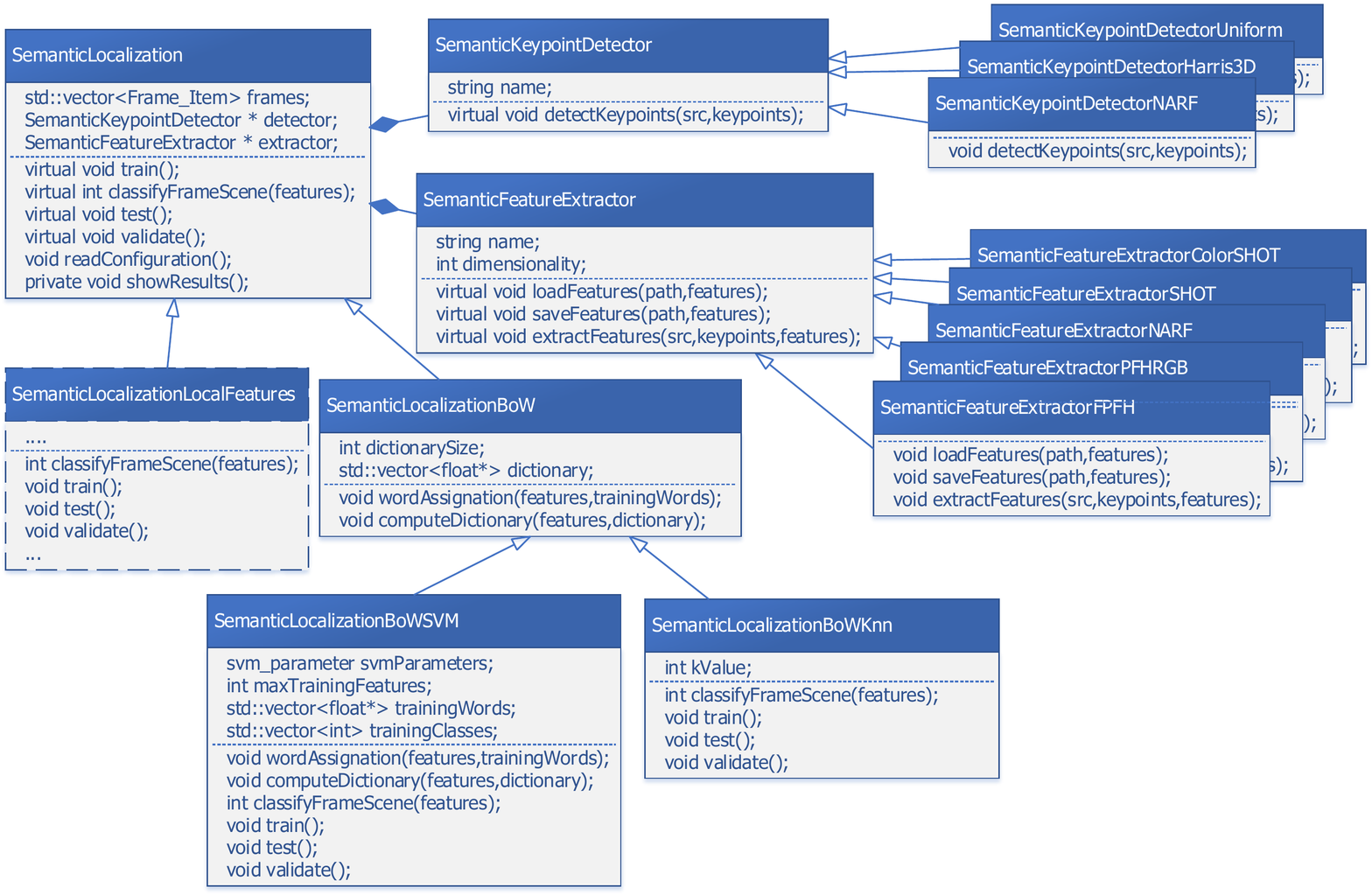}
\caption{Class diagram of the implemented framework.}
\label{classDiagram}
\end{figure}

For that reason, we have defined a diagram class (see Fig.~\ref{classDiagram}) where several abstract classes and methods are presented. The \texttt{SemanticLocalization} class implements some methods: \texttt{readConfiguration}, which reads a configuration file containing the point clouds to be used as input to the method; \texttt{test} and \texttt{validate} used for testing and validating the method (these methods call the \texttt{train} and \texttt{classify} abstract methods and, finally, \texttt{showResults} which shows the results of the classification. So \texttt{train} and \texttt{classifyFrame} are abstract and must be implemented in inherited classes. This class also has several attributes: \texttt{frames} are the point clouds to use in the classification and \texttt{detector} and \texttt{features} are the keypoint detector method and feature descriptor to be used in the classification, respectively.

We also provide two different classification methods, both making use of the BoW descriptors as input data. The first one is the Support Vector Machine (SVM) \cite{burges1998tutorial}, which learns to classify elements from two different classes finding a hyperplane which provides less classification error. By other hand, we have used the $k$-Nearest-Neighbors ($k-$NN)~\cite{cover1967nearest} method that directly uses the training data as model. Given a new element to classify, the $k$ nearest neighbors from the training data are selected. The new element is assigned to the class with more elements in the neighborhood. Other supervised classification methods could be incorporated easily. 

The \texttt{SemanticLocalizationBoW} class inherits from \texttt{SemanticLocalization} and uses a BoW approach. To do that, an attribute class \texttt{dictionary} contains the dictionary to be used in the classification process. In this class, two methods are implemented: \texttt{computeDictionary} which must be called before training and \texttt{wordsAssignation} where the words from the data are calculated. From this class, two other classes are defined, depending on the classification method used: \texttt{SemanticLocalizationBoWSVM}, that needs to define a \texttt{SVMModel}, and \texttt{SemanticLocalizationBoWKNN} which does not need to define any additional attribute.

Using this scheme, the final user can focus on implementing its method, or using different keypoint detectors and feature descriptors, thus providing an easy way to make experiments in semantic localization.

\section{Experimental results\label{sec:experiments}}

\subsection{Dataset description: ViDRILO}

All the experimentation included in this article has been carried out using ViDRILO: the Visual and Depth Robot Indoor Localization with Objects information dataset\footnote{http://www.rovit.ua.es/dataset/vidrilo/}. This dataset, whose overall characteristics are shown in Table~\ref{tab:seqInfo}, provides five different sequences of RGB-D images captured by a mobile robot within an office indoor environment. 

\begin{table}[!htb]
 \centering
 \caption{Overall ViDRILO sequences distribution.}
	\begin{tabular}{@{\extracolsep\fill}lcccc  } 
	\hline
	Sequence & Number of Frames & Floors imaged & Dark Rooms & Time Span \\   \hline
    Sequence 1   & 2389 	&  1st,2nd & 0/18 & 0 months \\
    Sequence 2   & 4579 	&  1st,2nd & 0/18 & 0 months \\
    Sequence 3   & 2248 	&  2nd & 4/13 & 3 months \\
    Sequence 4   & 4826 	&  1st,2nd & 6/18 & 6 months \\
    Sequence 5   & 8412 	&  1st,2nd & 0/20 & 12 months \\ 
    \hline
    \end{tabular}
 \label{tab:seqInfo}
\end{table}

\begin{figure}[!tbh]
 \begin{center}
 \begin{tabular}{@{\extracolsep\fill} c c c c c}
  \includegraphics[width=0.195\linewidth]{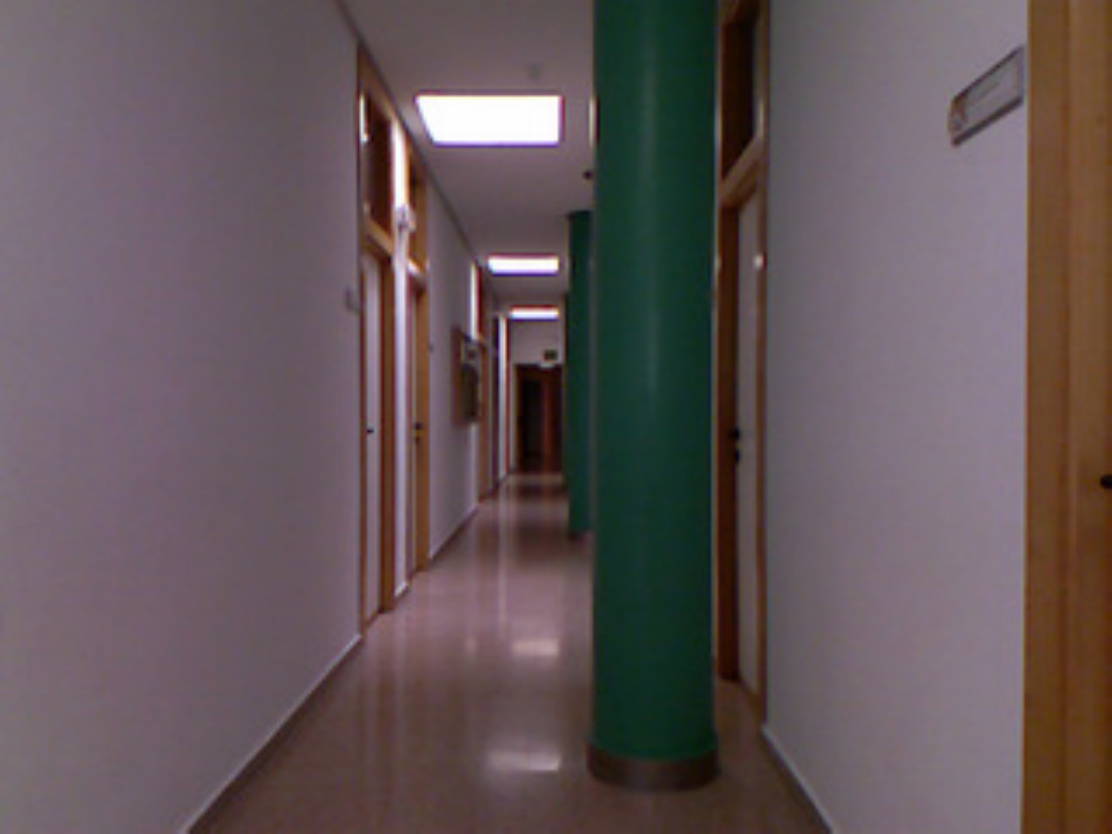}
  &
  \includegraphics[width=0.195\linewidth]{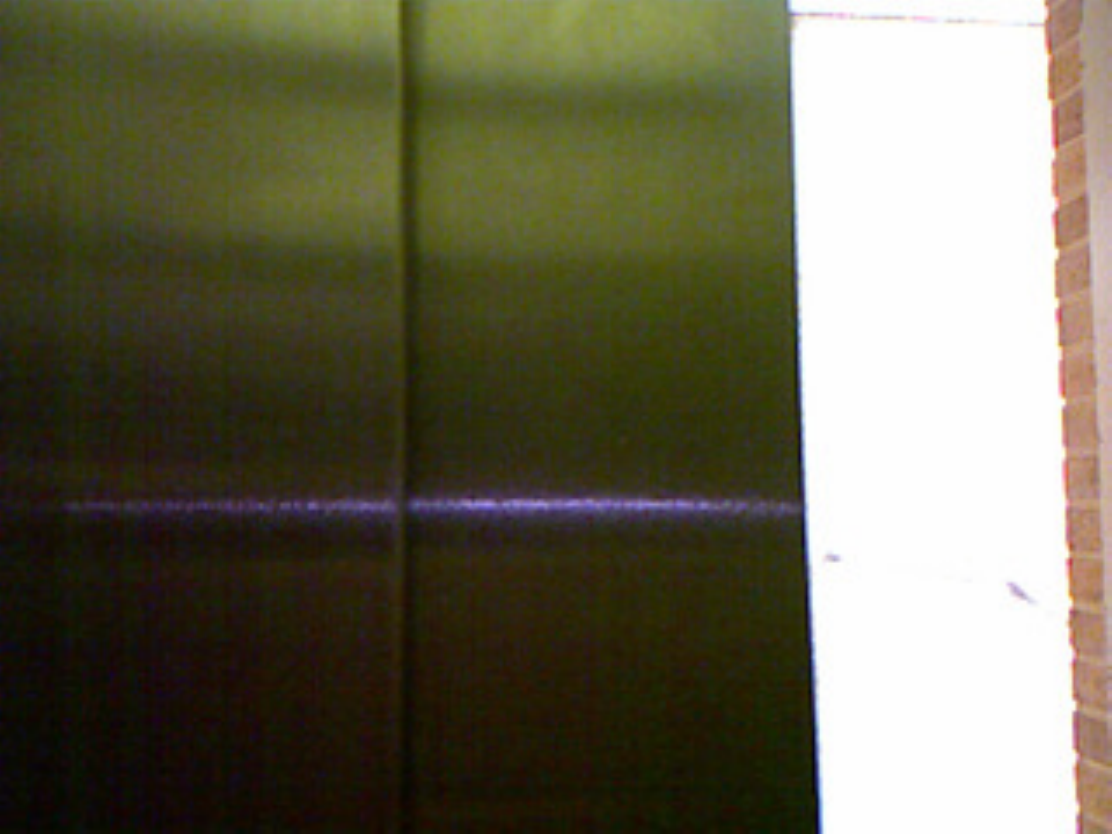}
  &
  \includegraphics[width=0.195\linewidth]{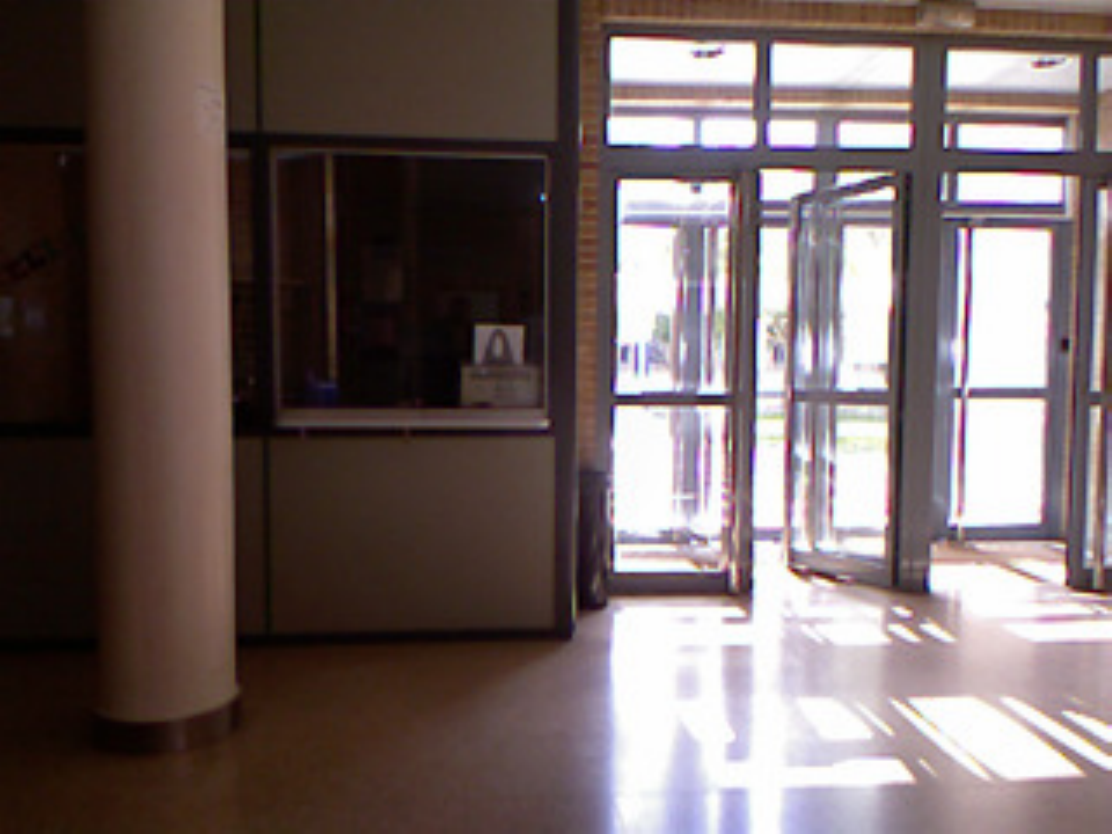}
  &
  \includegraphics[width=0.195\linewidth]{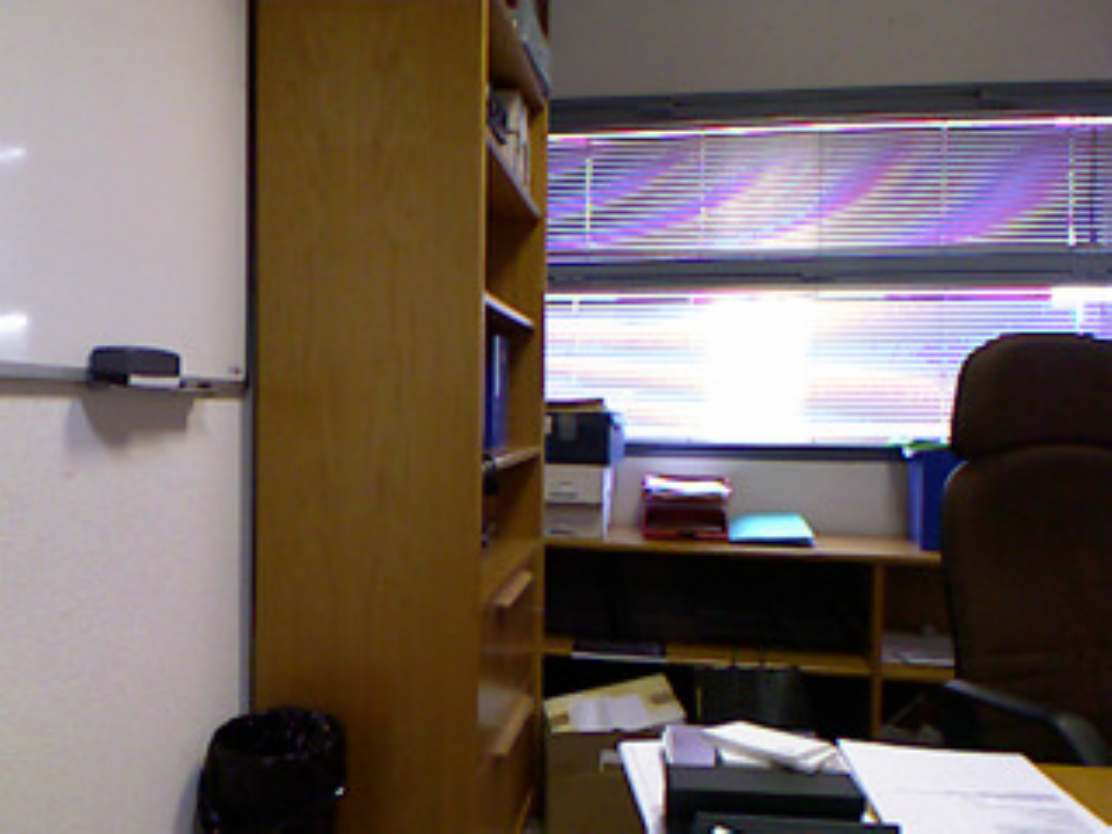}
  &
  \includegraphics[width=0.195\linewidth]{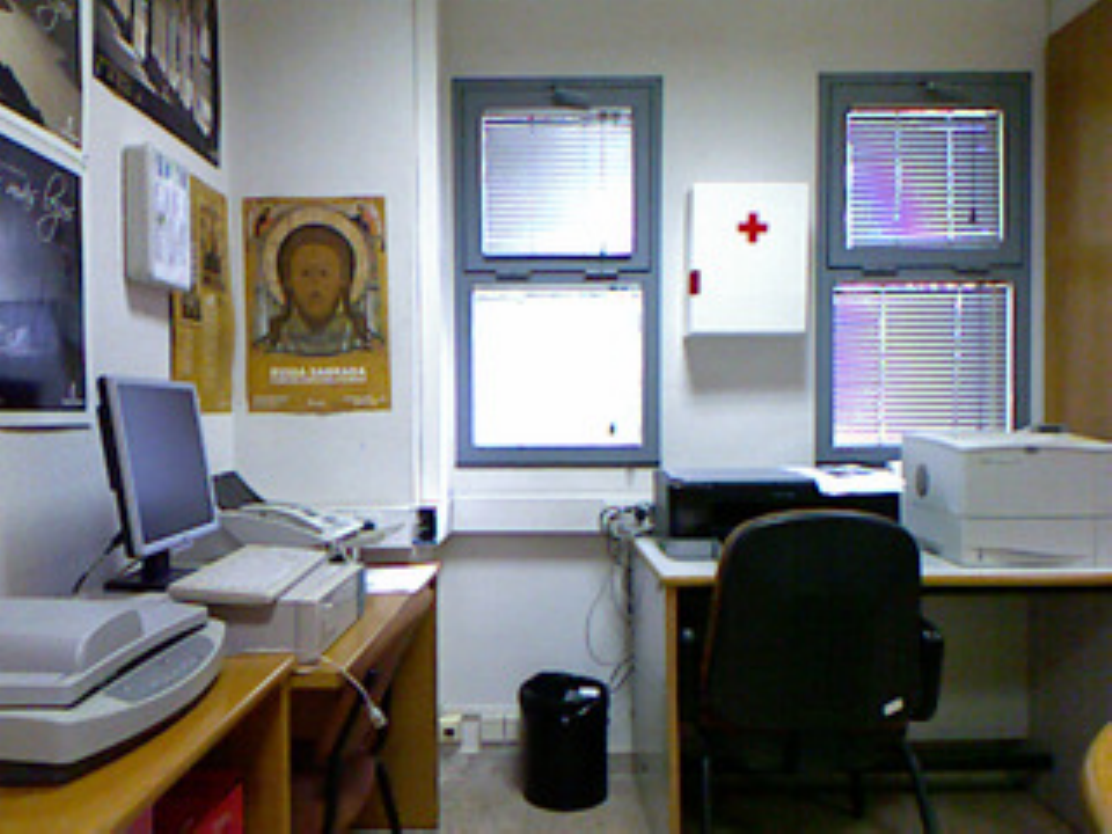}
   \\   
  CR &
  EA &
  HA &
  PO &
  SE
   \\
  \includegraphics[width=0.195\linewidth]{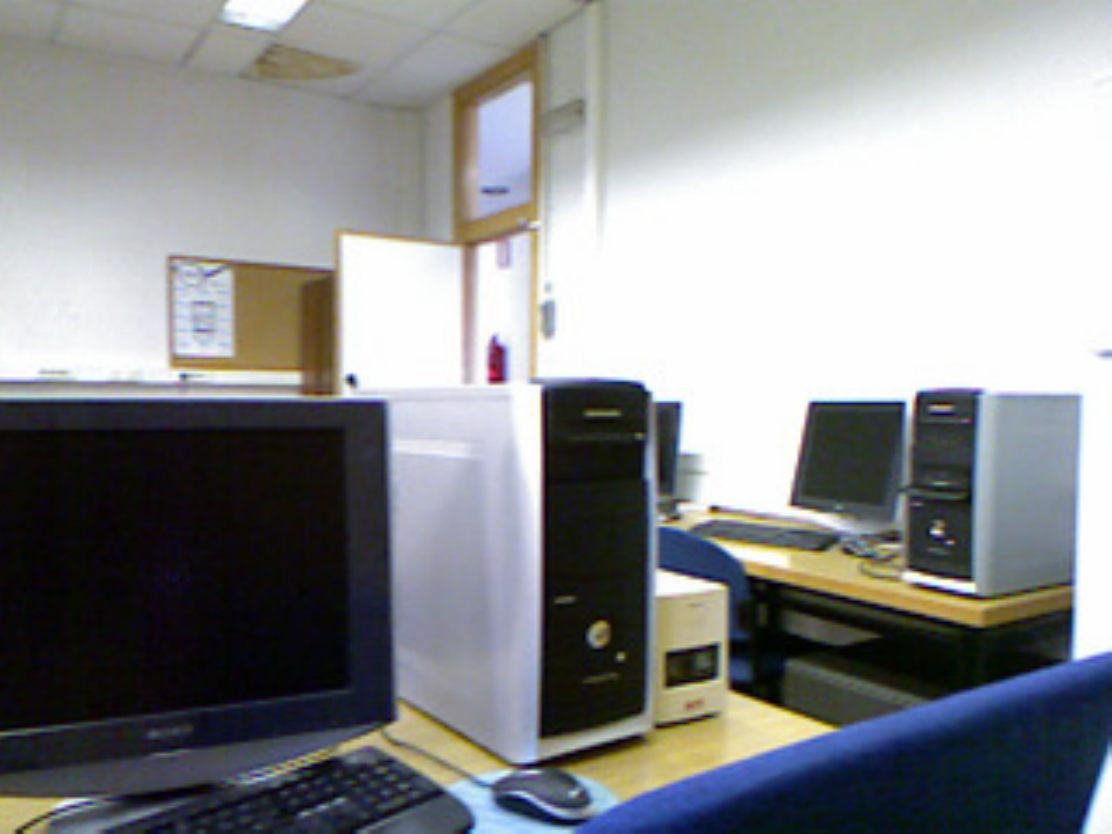}
  &
  \includegraphics[width=0.195\linewidth]{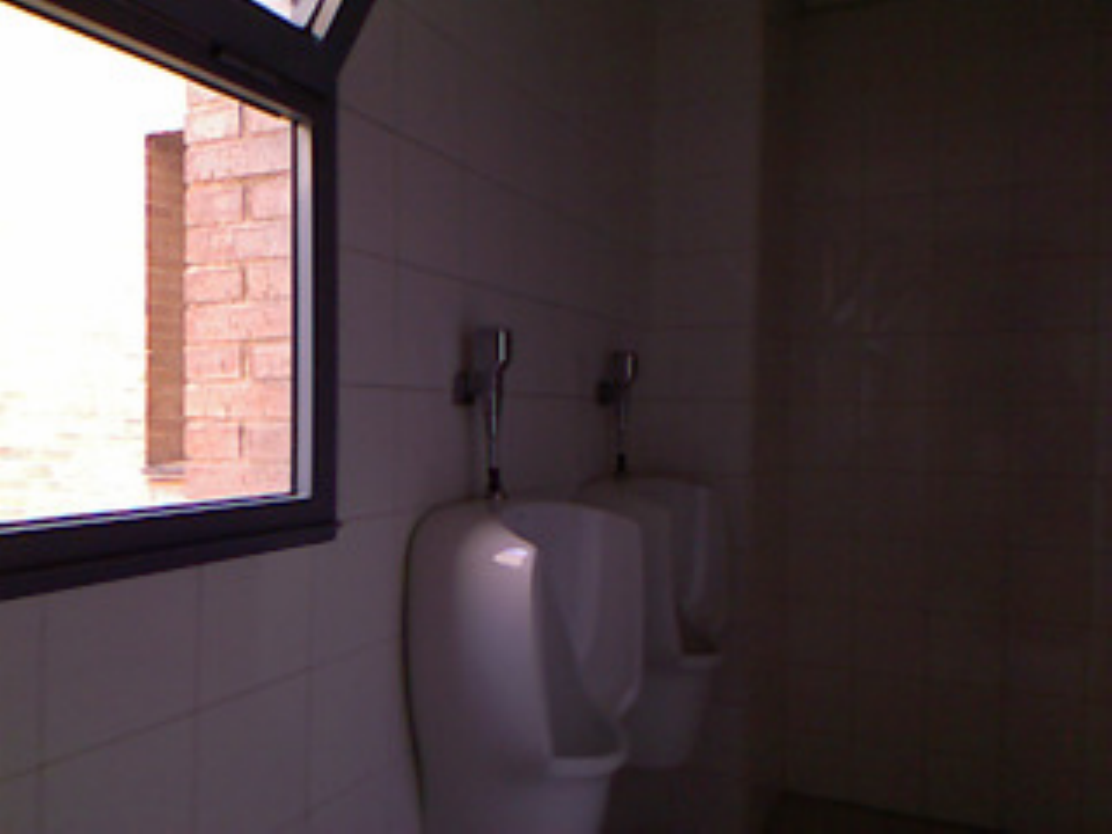}
  &
  \includegraphics[width=0.195\linewidth]{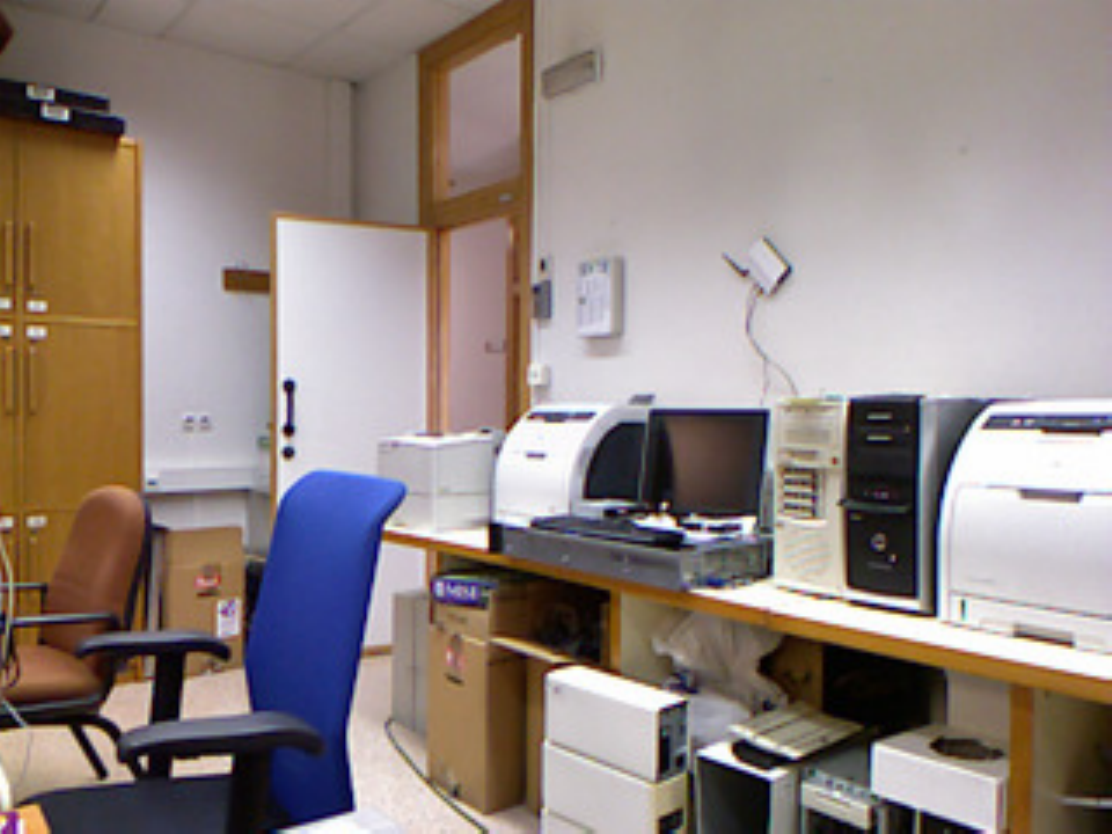}
  &
  \includegraphics[width=0.195\linewidth]{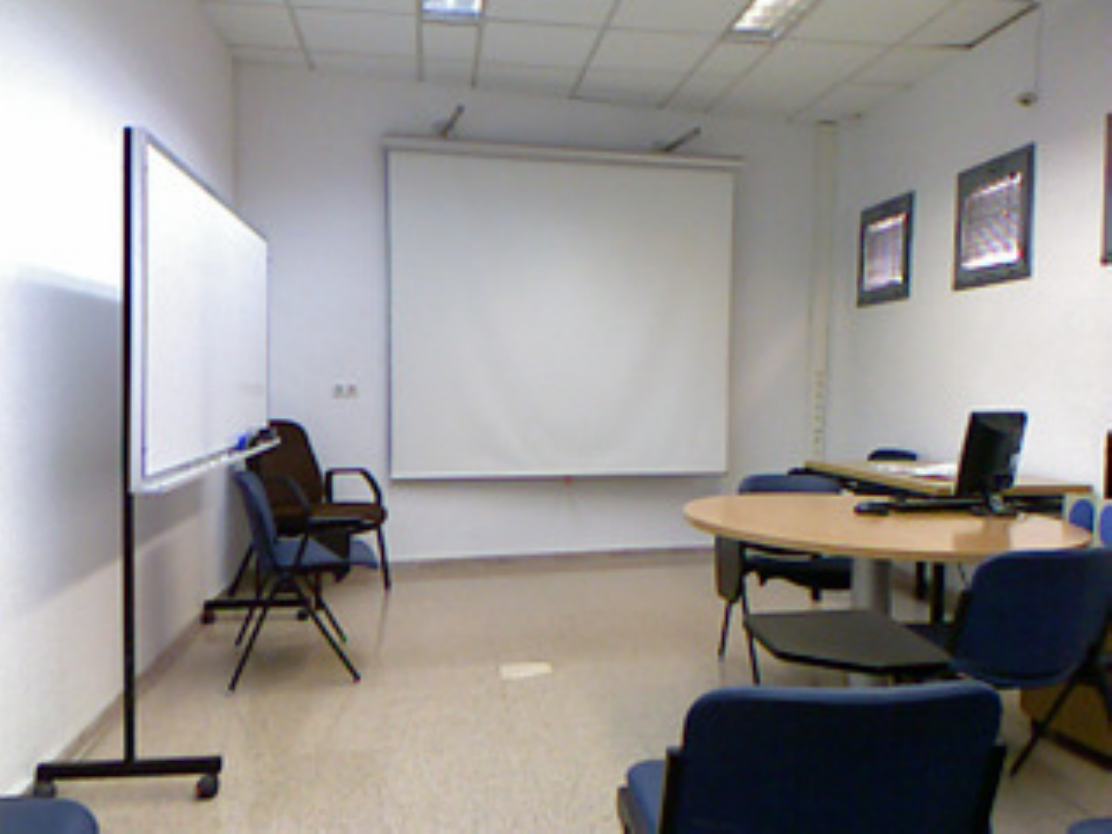}
  &
  \includegraphics[width=0.195\linewidth]{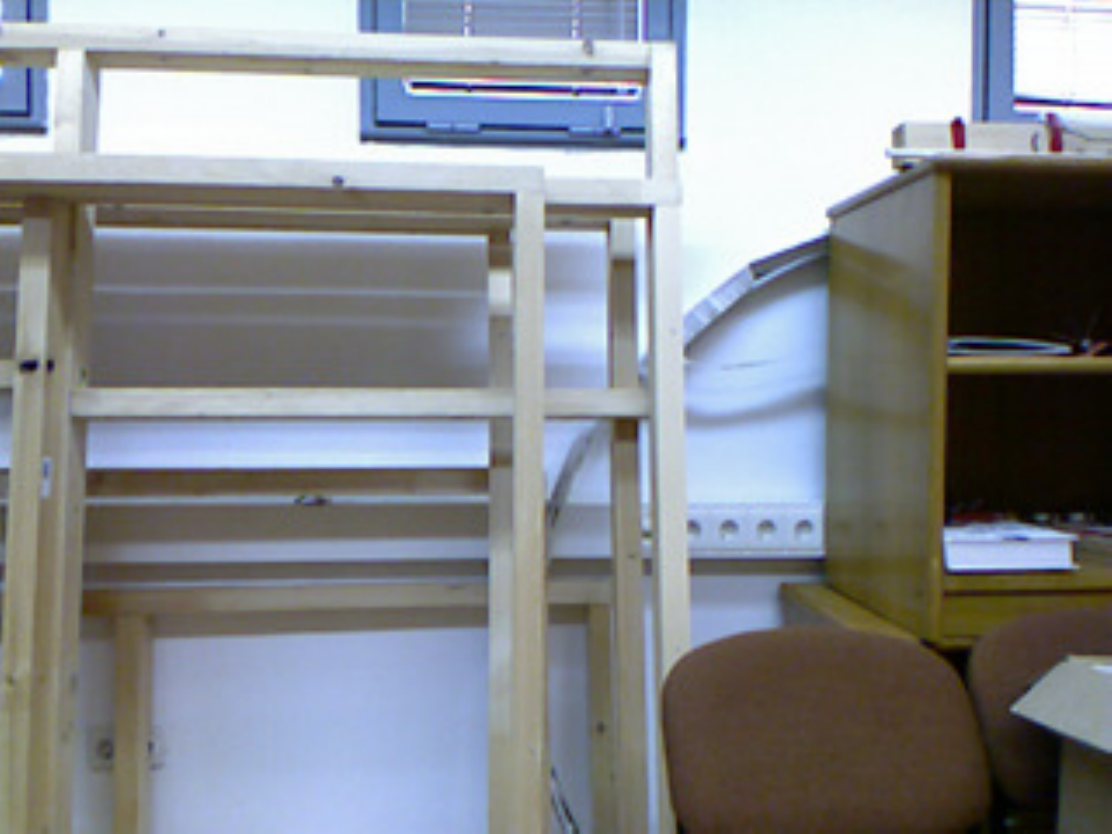} 
  \\
  SO &
  TO &
  TR &
  VC &
  WH  
  
  \end{tabular} 
  \end{center}
  \caption{Exemplar visual images for all room categories in ViDRILO.}
 \label{fig:fig2} 
 \end{figure}

Each RGB-D image is annotated with the semantic category of the room it was acquired, from a set of ten room categories. Unreleased sequences from ViDRILO have been successfully used in the RobotVision at ImageCLEF competition~\cite{martinez2012overview} in 2013~\cite{caputo2013imageclef} and 2014~\cite{caputo2014imageclef}. Fig.~\ref{fig:fig2} shows exemplar images for each one of the ten room categories using the following codes: CR (Corridor), HA (Hall), PO (Professor Office), SO (Student Office), TR (Technical Room), TO (Toilet), SE (Secretary Office), VC (Video Conference Room), WH (Warehouse), and EA (Elevator Area).
\setlength{\tabcolsep}{0.1em}

\begin{figure}[!tbh]
 \centering
 \includegraphics[width=0.8\linewidth]{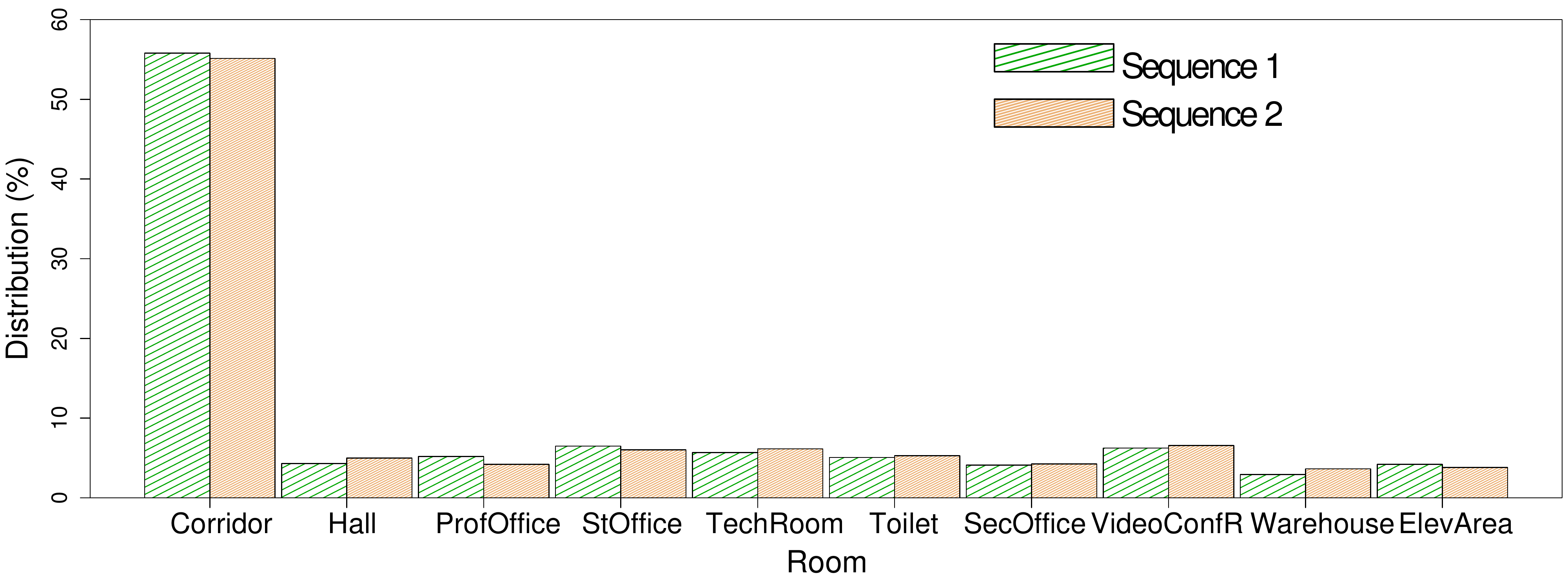}
 \caption{Room distribution for Sequences 1 and 2 in the ViDRILO dataset.}
 \label{fig:fig3}
\end{figure}

To focus on the internal parameters of the BoW approach, the experimentation stage is limited to the use of Sequence 1 and Sequence 2 from the dataset. The room distribution for these sequences is shown in Fig.~\ref{fig:fig3}. Here, we can observe that we are facing a challenging problem due to the dataset is highly unbalanced: most of the RGB-D images belong to the "Corridor" category.

\subsection{Study of keypoints detection}

Three different keypoints detection methods are evaluated in this work: NARF, Harris3D and Uniform Sampling, all they implemented in the PCL. These methods select a subset of 3D points from an input cloud using different methods, but they differ in the average amount of selected points. In the following, we describe the internal parameters used for the experimentation. We only fixed those parameters that should be explicitly established. The rest of parameters were set to their default values. Regarding the NARF detector, we used a support size of 20 cm. This parameter represents the diameter of the sphere used to find neighboring points, and therefore to estimate if a point belongs to a border or not. 
With respect to the Harris3D detector, we have used a threshold of 0.01 as we found it as a reasonable value to remove weak keypoints. Finally, the Uniform Sampling detector internally uses a voxel grid unsupervised downsampling method. We opted to use a radius of 0.03 m, which means we get a representative point each $0.03m^3$ area. 

\begin{figure}[!tbh]
 \centering
 \begin{tabular}{c c c}
 
 \multicolumn{3}{c}{
 \includegraphics[width=0.5\linewidth]{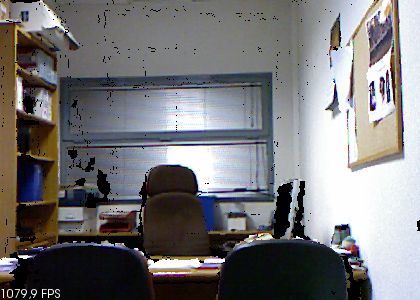}}
  \\
  \multicolumn{3}{c}{Input RGB-D Image}
  \\
  
 \includegraphics[width=0.3\linewidth]{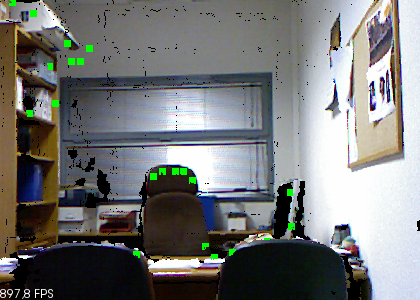} & 
 \includegraphics[width=0.3\linewidth]{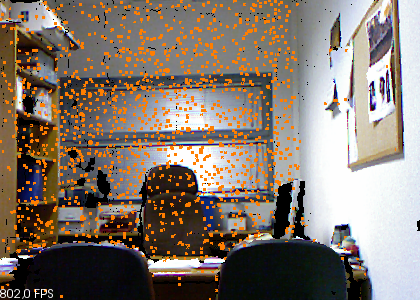} & 
 \includegraphics[width=0.3\linewidth]{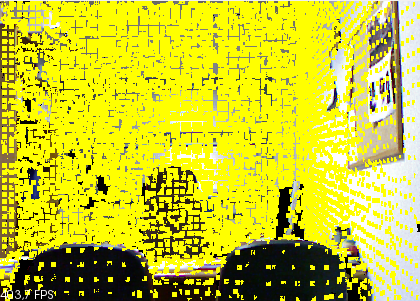} 
 \\ 
 NARF: 27 &  Harris3D: 2445 & Unif. Sampling: 12412 
 \end{tabular}
 
 \caption{Keypoint detection with NARF (bottom left), Harris3D (bottom center) and Uniform Sampling (botoom right) for a sample RGB-D image (top). The number indicates the amount of keypoints detected with each method}
 \label{fig:fig4}
\end{figure}

Fig.~\ref{fig:fig4} graphically presents the keypoint detection with these three techniques. We selected NARF, Harris3D and Uniform Sampling to study the effect of detecting a small, medium and large number of keypoints respectively. 

\subsection{Semantic Localization results}

We test our approach for the generation of semantic localization systems on the ViDRILO dataset. Concretely, we evaluated the generalization capabilities by generating classifiers using Sequence 2 (2479 RGB-D images) for training. These systems are then used to classify the 2389 RGB-D images from Sequence 1. Both sequences were acquired in the same building during two consecutive days. The robot used for the acquisition followed a similar path but in the opposite direction, which affects the viewpoint of the imaged scenes. The following internal parameters are evaluated:

\begin{itemize}
	\item 3 Keypoint detectors: NARF, Harris3D and Uniform Sampling.
    \item 5 Feature extractors: NARF, SHOT, Color-SHOT, PFH-RGB, and FPFH.
    \item 4 Dictionary sizes: 25, 50, 100 and 200.
    \item 2 Classification models: 
    \begin{itemize}
    	\item SVM classifier (exponential chi-square kernel).
        \item $k$-Nearest-Neighbor ($k=7$).
    \end{itemize}
\end{itemize}

\begin{figure}[!tbh]
 \centering
 \begin{tabular}{c c}
 
 \multicolumn{2}{c}{Keypoint Detection = NARF}
 \\ 
 SVM Classifier & $k-$NN Classifier 
 \\ 
\includegraphics[width=0.45\linewidth]{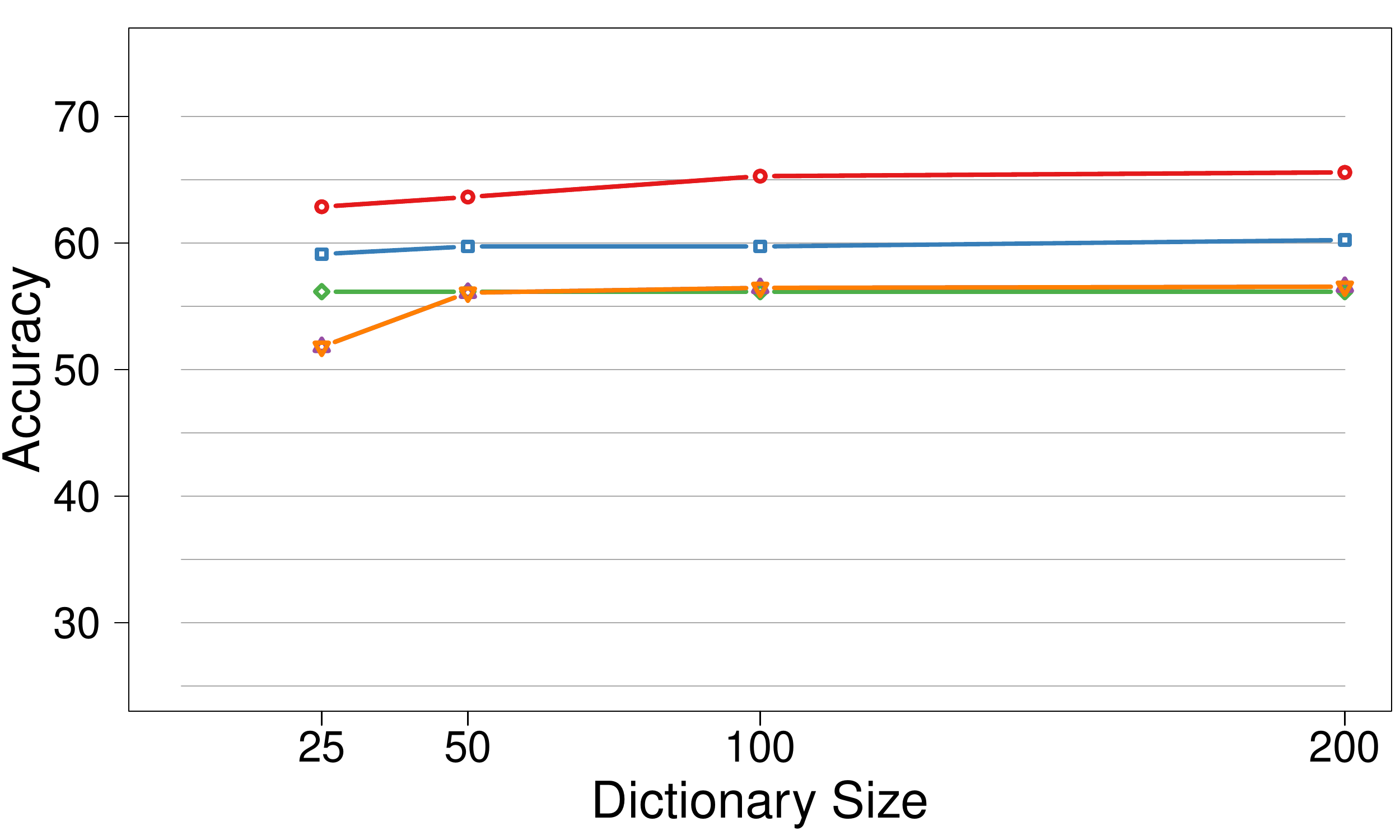} 
& 
 \includegraphics[width=0.45\linewidth]{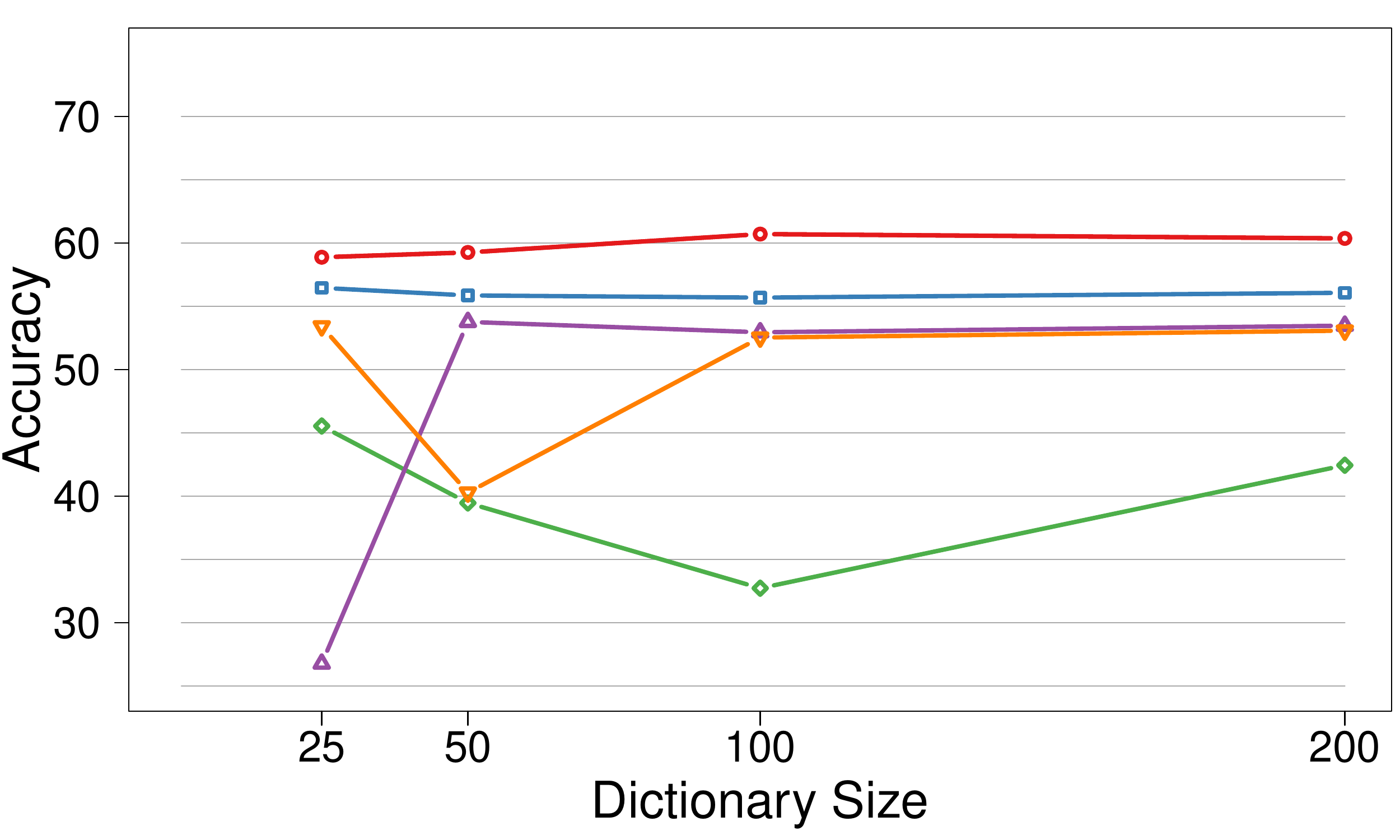} 
 \\
 \multicolumn{2}{c}{\includegraphics[width=0.85\linewidth]{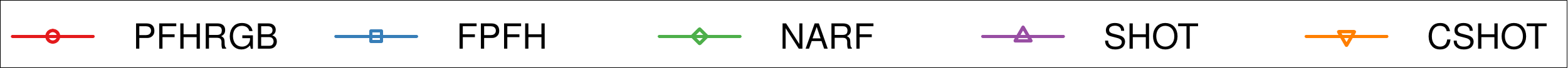} }
 \\
 \multicolumn{2}{c}{Keypoint Detection = Harris3D}
 \\ 
 SVM Classifier & $k$-NN Classifier 
 \\ 
 \includegraphics[width=0.45\linewidth]{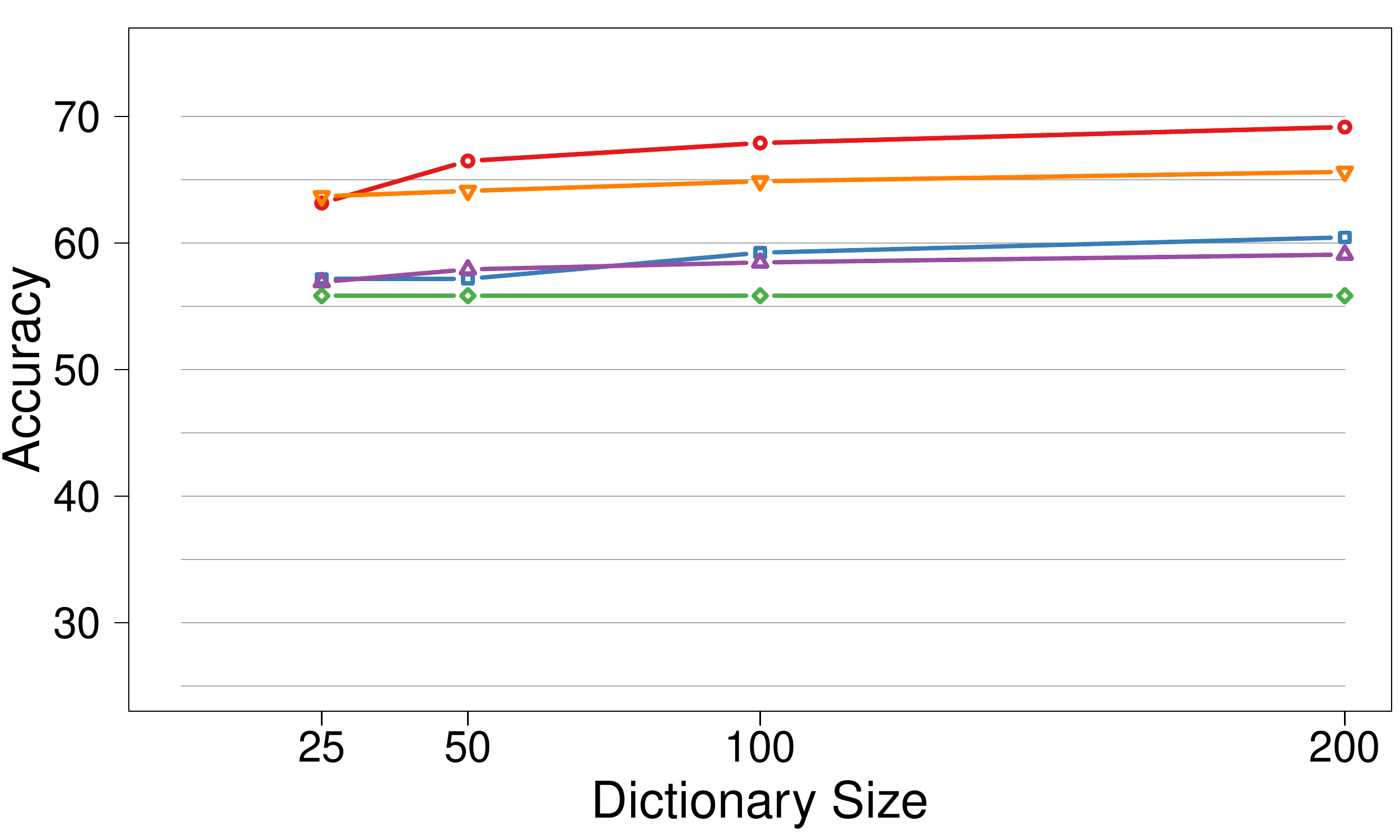} 
 &
 \includegraphics[width=0.45\linewidth]{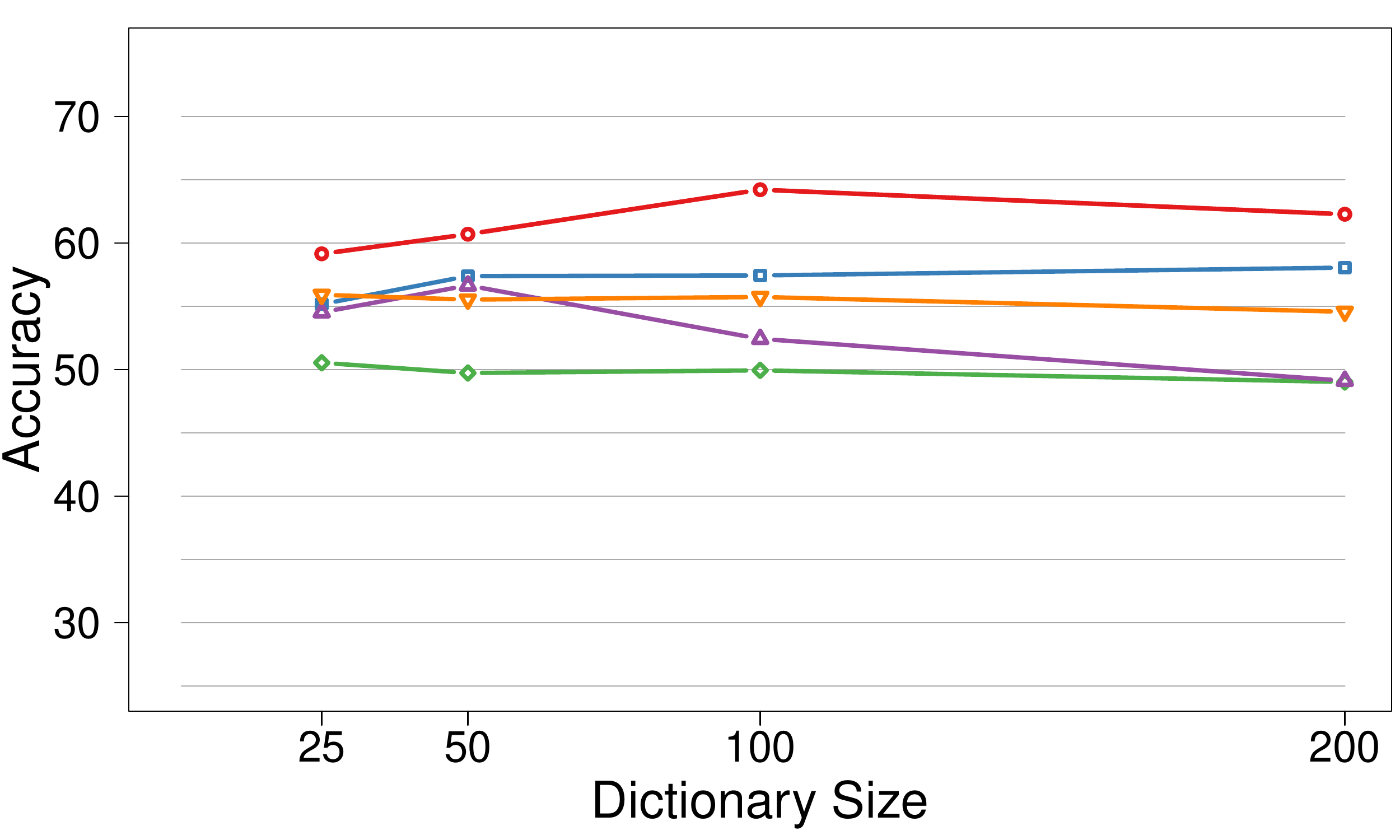} 
 \\
 \multicolumn{2}{c}{\includegraphics[width=0.85\linewidth]{legend.pdf} }
 \\
 \multicolumn{2}{c}{Keypoint Detection = Uniform Sampling}
 \\ 
 SVM Classifier & $k$-NN Classifier 
 \\ 
 \includegraphics[width=0.45\linewidth]{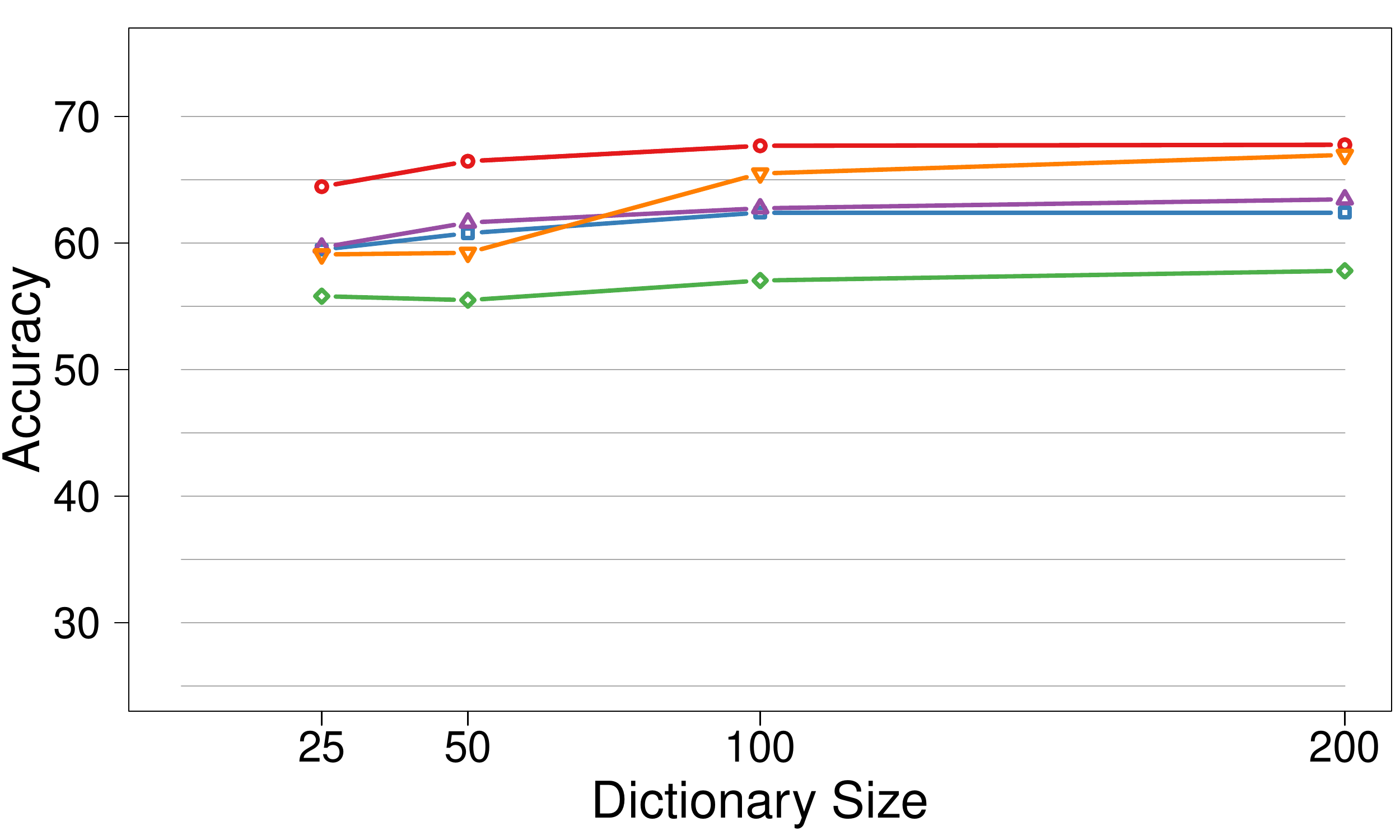} 
 &
 \includegraphics[width=0.45\linewidth]{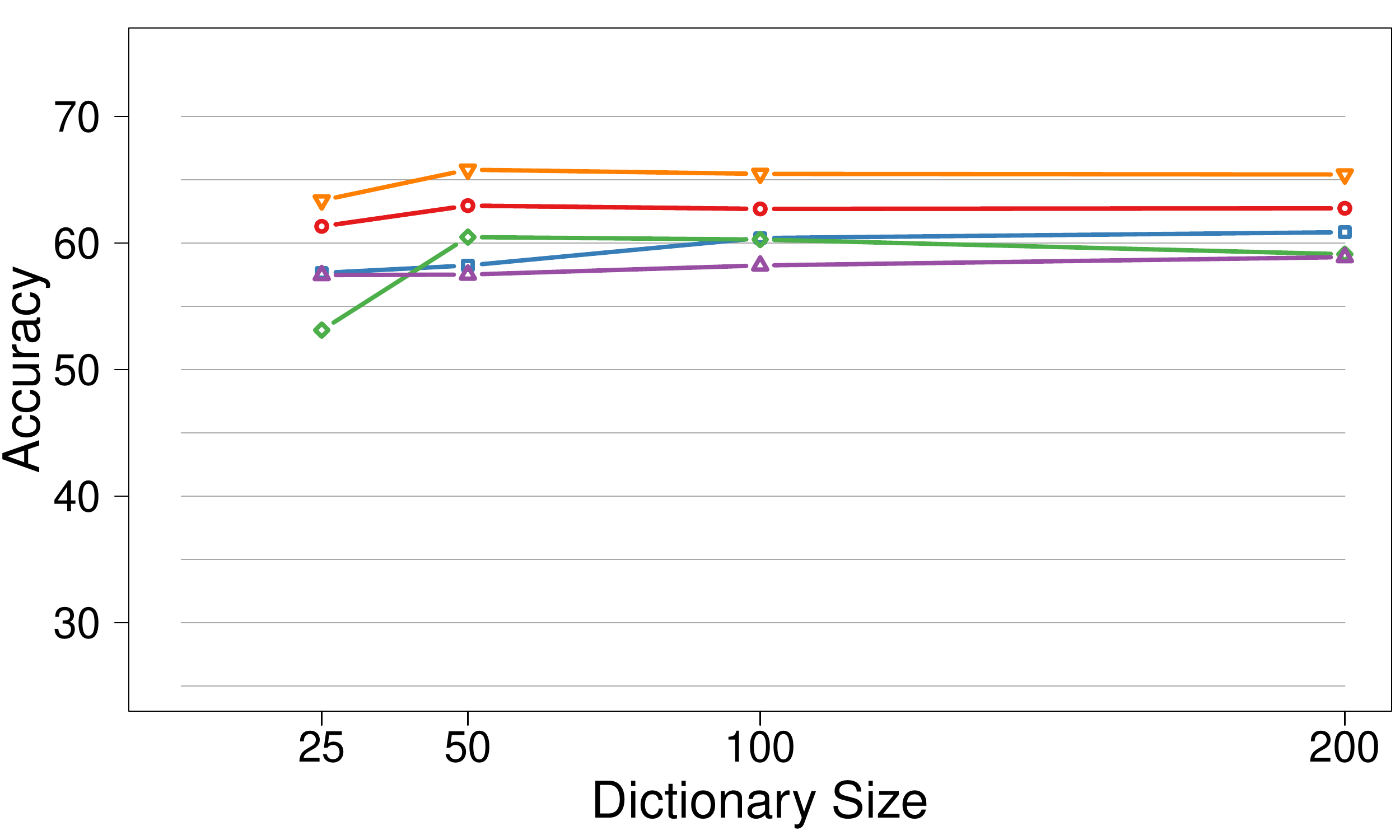} 
 \\
 \multicolumn{2}{c}{\includegraphics[width=0.85\linewidth]{legend.pdf} }
 
 \end{tabular}
 
 \caption{Semantic localization overall results. Accuracy values obtained by training a SVM (left) or $k$-NN classifier (right) with Sequence 2 and evaluating against Sequence 1 from the ViDRILO dataset.}
 \label{fig:fig5}
\end{figure}

Fig.~\ref{fig:fig5} shows the accuracies obtained with all the semantic classifiers, and we can extract some remarks from these results. Firstly, we can observe that the SVM classifier outperforms the use of $k$-NN in most of the cases. The two classification models evaluated in this work behave different with respect to the dictionary size. Increasing the size of the dictionary always has a positive impact on the accuracy when using SVM, but not with $k$-NN. Regarding the keypoint detection method, NARF is the one presenting the worst results, as it could have been expected. At this point, we should outline the bad behavior of the combination of NARF as keypoint detector and feature extraction techniques. The main differences between Harris3D and Uniform Sampling are related to the classification models. That is, the improvement obtained thanks to the use of Uniform Sampling (with respects to Harris3D) is notoriously greater when using $k$-NN as classification model.

An analysis of the feature extraction methods exposes PFHRGB and Color-SHOT as the most promising techniques. On the contrary, NARF, FPFG and SHOT features present the lower accuracies. It should be taken into account that PFHRGB and Color-SHOT are the only two features that integrate color information. The overall highest accuracy (69.17) was obtained with a SVM and a combination of Harris3D and PFHRGB as keypoint detector and feature extractor respectively. Therefore, we can conclude that the use of Uniform Sampling is not needed unless a $k$-NN classifier is used. The use of Harris3D as keypoint detection technique notoriously reduces the amount of data to work with and speeds up the 3D processing. 

We also evaluated the use of one of the state-of-the-art global 3D feature: the Ensemble of Shape Functions (ESF)~\cite{wohlkinger2011ensemble}. Using the ESF descriptor, we trained both SVM and $k$-NN classifiers from Sequence 2 and tested against Sequence 1. We obtained an accuracy value of 58.48\% with $k$-NN and 64.49\% with the SVM classifier. 
Consequently, the BoW approach allowed us to outperform the ESF global descriptor. Moreover, we obtained better results using descriptors whose dimensionality is notoriously lower than for the ESF descriptor (200 vs 640). This difference in the descriptor dimensionality would result in classification models that can be trained in a lower amount of time, and perform RGB-D images classification much faster. 

\section{Conclusions and future work\label{sec:conclussions}}

Semantic localization is a challenging problem in robotics. We have presented in this article a framework for the generation of global 3D descriptors from local ones following a BoW approach. This framework has been implemented in the Point Cloud Library and evaluated in the semantic localization problem. 

Based on the experimentation stage, we can affirm that PFHRGB and Color-SHOT are the two 3D local features with the best performance. Harris3D exposed as the most appropriate keypoint detection method, due to it notoriously reduces the amount of data to work with respects to Uniform Sampling. The proposed BoW framework obtained higher accuracies that the use of the well-known global 3D feature ESF.

As future work, we have in mind the experimentation with a wider variety of 3D features and keypoint detection methods. Moreover, larger dictionary sizes will also be considered.

\section*{Acknowledgments}
\label{sec:ack}

This work was supported by grant DPI2013-40534-R of the Ministerio de Economia y Competitividad of the Spanish Government, and by Consejer\'ia de Educaci\'on, Cultura y Deportes of the JCCM regional government through project PPII-2014-015-P. Jesus Mart\'inez-G\'omez is also funded by the JCCM grant POST2014/8171.

\bibliographystyle{apalike}
{\small
\bibliography{references}}

\vfill

\end{document}